\documentclass[11pt,a4paper]{article}
\usepackage[T1]{fontenc}
\usepackage[utf8]{inputenc}
\usepackage{lmodern}
\usepackage[margin=25mm]{geometry}
\usepackage{amsmath,amssymb,mathtools}
\usepackage{graphicx,booktabs,tabularx,longtable,array,microtype}
\usepackage[numbers,sort&compress]{natbib}
\usepackage{xcolor,enumitem,url}
\usepackage[colorlinks=true,linkcolor=teal,citecolor=teal,urlcolor=teal]{hyperref}
\usepackage{caption,fancyhdr,placeins,tikz}
\usetikzlibrary{arrows.meta,positioning}
\hypersetup{pdftitle={Selective Monocular Precession Estimation under Geometric and Evaluation Ambiguity},pdfauthor={Carlos Eduardo Sanoja}}
\setlist{itemsep=2pt,topsep=4pt}
\newcolumntype{Y}{>{\raggedright\arraybackslash}X}
\newcommand{\RP}{\mathbb{RP}^{2}}
\newcommand{\SO}{\mathrm{SO}(3)}
\newcommand{\R}{\mathbb{R}}

\newcommand{\method}{ORBIT-SCOUT}
\newcommand{\hash}[1]{{\footnotesize\path{#1}}}
\title{\textbf{Selective Monocular Precession Estimation\\under Geometric and Evaluation Ambiguity}\\[7pt]
 \large An Auditable Synthetic Study with ORBIT-SCOUT}
\author{Carlos Eduardo Sanoja\\[4pt]
 {\small Quanta Labs, LLC}\\
 {\small Universidad Monte\'avila}\\
 {\small\href{https://orcid.org/0009-0000-0339-7072}{ORCID: 0009-0000-0339-7072}}}
\date{25 October 2026}
\begin{document}
\maketitle
\begin{abstract}
Monocular observations of a tumbling object can support a precise fitted trajectory without establishing that the recovered physical axis is correct. We study this distinction for elongated, approximately cylindrical targets observed by a calibrated stationary camera. ORBIT-SCOUT couples image-derived line and rim geometry to constant-cone and torque-free triaxial trajectory families, then returns a linked momentum-axis and mean-precession-rate estimate only when numerical, geometric, temporal and competing-hypothesis checks pass. We describe the measurement objective, selection policy and limitations, and situate them within spacecraft pose estimation, unknown-target reconstruction, rigid-body identification and selective prediction. An internally preregistered population of 96 new synthetic scenes was processed in two cold passes, retaining all 192 attempts. In the primary pass, 38 scenes produced jointly correct estimates, none produced an incorrect estimate, 57 were withheld and one execution was incomplete. Correct coverage on the 64 supported scenes was 25/64. Ninety-five repeated pairs were scientifically equivalent under a frozen comparison policy; the remaining pair was incomplete. The registered coverage, failure-free execution and all-case repeatability targets were not met. Historical audits also exposed an overly permissive axis metric and motion-dependent acquisition cadence. The contribution is a documented measurement-to-decision implementation and a failure-preserving evaluation case study, not a new rigid-body theory or evidence of state-of-the-art accuracy. Zero observed errors does not establish calibrated risk, and conditional temporal checks do not remove shared image-feature bias. Real-image transfer and external reproduction remain open.
\end{abstract}

\section{Introduction}
The practical question in observing a tumbling object is not merely whether a model can be fitted. It is whether the available images support the physical quantity a downstream system intends to use. A projected line can belong to a cylinder generator, an end rim or surface texture. A regular sequence can admit competing frequencies. A repeatable implementation can preserve the same incorrect geometric interpretation. These issues become consequential when an estimator reports a narrow uncertainty measure or an apparently perfect success count.

This paper examines a restricted problem: estimating an unoriented angular-momentum direction in camera coordinates and a positive mean precession rate from one calibrated video. The target is assumed to be elongated and approximately cylindrical, its optical longitudinal axis is assumed to align with the relevant principal inertia axis, and the camera is stationary. We do not estimate complete six-degree-of-freedom pose, arbitrary translational motion, absolute inertia, a capture trajectory or all components of angular velocity. These distinctions are essential when comparing the problem with spacecraft navigation and pose estimation.

The literature already contains model-based monocular pose initialization, learned spacecraft pose networks, methods for unknown-target shape and motion, and inference of inertial properties from rotational observations \citep{sharma2018,sharma2020,capuano2020,chi2022,kobayashi2026}. Circular features and dynamical filtering have also been combined for non-cooperative targets \citep{zhang2023}. Consequently, using one camera, avoiding an individual CAD input, fitting torque-free dynamics, retaining hypotheses or allowing rejection is not, by itself, a defensible novelty claim.

Our motivation instead comes from the interaction between physical interpretation and evaluation. During development, allowing a global depth reflection in an axis metric made two wrong solutions appear correct. A separate synthetic population tied image cadence to target rate, exposing a nonvisual proxy for a quantity intended to be inferred from images. Image-domain probes further showed that temporally coherent wrong edges can fit a motion model well. These observations prompted stricter accounting, an explicitly selective reporting contract and a fresh synthetic study. The resulting experiment is useful partly because it failed the planned acceptance targets.

The study asks three questions. First, what can a geometry-based trajectory fit legitimately certify under its stated measurement assumptions? Second, what joint accuracy and coverage does a frozen selective pipeline attain on new synthetic scenes when withholding and execution failure remain distinct? Third, which conclusions survive correction of the metric, acquisition design and repetition accounting? These are case-study questions. They do not assume that our pipeline is the strongest available estimator or that a small synthetic evaluation is sufficient for mission qualification.

The paper makes three bounded contributions:
\begin{enumerate}
 \item A documented implementation connecting calibrated endpoint geometry, two physical motion families, a fixed training-only selection rule and an explicit decision contract separating estimates, withholding and incomplete execution. An appendix derives the endpoint correction used in the objective; its mathematical ingredients are established geometry, not claimed new theory.
 \item A prospective synthetic evaluation with a fixed 96-scene denominator, two cold executions per scene, linked axis/rate scoring and preservation of incomplete attempts. The reported outcome is a failed registered validation with a separately authenticated descriptive score.
 \item An analysis of how axis equivalence, acquisition metadata, shared preprocessing and post-development reuse change the meaning of an apparent success. Historical corrections and negative numerical controls are retained rather than folded into the final accuracy count.
\end{enumerate}
The strongest empirical statement is narrow: the frozen method emitted 38 correct joint estimates in this particular 96-scene primary pass, with no observed incorrect emission and substantial abstention. The manuscript does not convert that observation into a probability of safety, a universal identifiability claim or a ranking against methods evaluated on different observables.

\section{Related work and the scope of comparison}\label{sec:related}
\subsection{Review procedure}
We conducted a structured narrative review with a literature cutoff of 25 October 2026. Searches combined terms for monocular spacecraft pose, unknown non-cooperative targets, rotational-state and inertia estimation, circular features, temporal tracking, uncertainty, selective prediction and evaluation leakage. Primary papers, author-hosted manuscripts, publisher records and official dataset descriptions were preferred. The accompanying search ledger records thematic queries, source locations and depth of inspection. Recent preprints are identified as such; a future conference date in a preprint is not treated as evidence of completed publication.

This is an extended problem-oriented review, not an exhaustive systematic review. There is no claim of a complete database census, duplicate-screening protocol or meta-analytic performance estimate. Some publisher full texts were unavailable; claims about those papers are limited to their accessible author abstracts and previews. In particular, we do not infer undocumented implementation choices from a title. The comparison axes are the observed quantity, target prior, sensor information, temporal assumptions and evidence domain. Numerical scores are not pooled across incompatible tasks.

Broader surveys already organize the field by sensing and prior target knowledge. Opromolla et al.\ \citep{opromolla2017} review electro-optical methods for cooperative and uncooperative targets, including known and unknown objects. Pauly et al.\ \citep{pauly2023} distinguish hybrid and direct learned pipelines and discuss datasets, computational deployment and the synthetic-to-real gap. Our review selects work relevant to the axis/rate estimand and evaluation integrity; these broader classifications and deployment concerns are established context.

\subsection{Known-target monocular pose estimation}
Classical model-based methods establish an important baseline for what image geometry can achieve when the object model is known. Sharma, Ventura and D'Amico \citep{sharma2018} use a wireframe model, extracted edges and geometric constraints to initialize pose from a single image, followed by refinement; their evaluation includes PRISMA flight imagery. Their feature-to-model correspondence problem differs from ours, but the work rules out treating edge-based monocular estimation or ambiguity-aware refinement as new contributions here.

The Spacecraft Pose Network \citep{sharma2020} replaces much of the hand-designed visual representation with a learned detector and attitude prediction, using geometric constraints to recover position. The target remains known. This distinction between \emph{non-cooperative} and \emph{unknown} is often obscured: an object may carry no fiducials or communication link while still having a detailed model available to the observer. Our method removes the individual CAD input but substitutes a restrictive cylindrical and dynamical prior. Neither setup is prior-free.

The Satellite Pose Estimation Challenge formalized shared data and metrics for comparing such systems \citep{kisantal2020}. That common-task structure is precisely what our study lacks for an external performance ranking. A pose error on a known Tango spacecraft is not the same endpoint as an unoriented momentum axis and a time-averaged precession rate on a cylindrical object. A method returning a full orientation may contain information from which our estimand could be derived, but doing so requires a defined temporal estimator, coordinate convention and reference trajectory. We have not performed that adaptation.

Photorealistic simulation supports training and controlled evaluation, but realism is not a binary property. Proen\c{c}a and Gao \citep{proenca2020} introduced URSO and a pose approach that represents orientation ambiguity with a mixture formulation. That work is relevant both to appearance variability and to the need to retain alternative attitudes. It does not imply that a low-dimensional geometry model is more reliable than a learned representation, or that either transfers without testing.

\subsection{Temporal pose tracking and dynamics}
Temporal estimation can reduce frame-level ambiguity while creating another failure mode: persistence of a wrong hypothesis. Park and D'Amico \citep{park2023ukf} combine learned pose observations with an adaptive unscented Kalman filter and evaluate the navigation pipeline on synthetic and robotic-testbed trajectories in SHIRT. Their method estimates relative orbital and attitude states of a known target. In contrast, our offline fit uses the full source video and conditional leave-block-out checks; it is not a causal navigation filter.

Recent work makes the handling of temporal alternatives explicit. HAT \citep{hat2026}, a September 2026 preprint, preserves competing CAD-based pose histories and uses inter-frame motion to anchor and fuse a monocular SLAM trajectory. It requires calibrated RGB images, metric CAD and target regions. Its retained histories are stronger precedent for temporal hypothesis management than a generic claim that our system keeps alternatives. Our bank is finite, offline and tied to a narrower motion family; its remaining rivals are a veto mechanism, not a global posterior.

TALON \citep{talon2026}, a May 2026 preprint, introduces lightweight temporal adapters to a frozen vision transformer and aligns patch tokens with keypoint structure. It illustrates that temporal information can be incorporated at the representation level rather than only after pose estimation. We have not tested learned temporal features, fine-tuned a foundation model or measured the cost of substituting such features in our pipeline. A comparison would need to preserve the same source videos and downstream estimand.

Taken together, these approaches show why using physics and time is too broad a contribution claim. The scientific choice is where temporal consistency enters: in image features, pose correspondence, state propagation, trajectory fitting or decision gates. Our checks operate after a shared image measurement stage. That location constrains what they can detect.

\subsection{Unknown-target reconstruction and navigation}
Unknown-target estimation predates recent foundation models. Capuano et al.\ \citep{capuano2020} describe simultaneous pose and shape estimation using optical flow and star-tracker information, together with a structure-from-motion route to a later model-based phase. This is a useful reminder that monocular imagery can be one component of a broader sensing and knowledge system. The auxiliary attitude information and navigation setting must be included in a fair comparison.

Park and D'Amico \citep{park2024shape} infer a compact assembly of superquadrics and pose from one image and introduce a multi-object spacecraft reconstruction dataset. Their target prior is learned across shapes rather than supplied as one exact CAD model. Legrand et al.\ \citep{legrand2024} instead construct a target-specific radiance field from a sparse image collection, then render training images for an off-the-shelf pose estimator. Their pipeline selects and annotates target views and evaluates on SPEED+ hardware-in-the-loop imagery. Thus, absence of an original CAD model can still require target-specific reconstruction and pose annotations.

FoundationPose \citep{foundationpose2024} provides a broader example of unified model-based and model-free object pose estimation: a CAD model or reference imagery supplies the object representation, and its described pose pipeline consumes RGB-D observations. It is not a directly executed monocular spacecraft-precession baseline here. DreamSat-Pose \citep{dreamsat2026}, a July 2026 preprint revised in September, combines single-view shape reconstruction with learned image-to-geometry correspondence and PnP. This further weakens any claim that estimating something about an unseen target from a single camera is itself novel.

Francesch Huc and D'Amico \citep{huc2026} propose transformer-aided monocular odometry within a multi-state constraint Kalman filter for previously unseen spacecraft. Their October 2026 preprint uses servicer attitude excitation to recover metric information. Its stated simulation assumptions include exact servicer knowledge and calibration, target pointing and reference-derived bounding boxes. Attitude-history evaluation permits one constant rotation because the target frame is arbitrary. These are material conditions, not reasons to dismiss the work. They demonstrate a wider navigation objective under different support assumptions; our fixed-camera, offline axis/rate experiment neither reproduces nor supersedes it.

The central lesson is to specify the information budget. One RGB camera does not determine how much is known about geometry, calibration, segmentation, servicer state, illumination or dynamics. A reduction in one kind of prior can be offset by another. Table~\ref{tab:comparison} makes these distinctions explicit.

\begin{table}[t]
\centering\small
\caption{Selected nearby approaches and information assumptions. This is a task comparison, not a performance ranking. Simulation does not imply identical rendering, noise or dynamics.}\label{tab:comparison}
\begin{tabularx}{\linewidth}{>{\raggedright\arraybackslash}p{27mm}YYY}
\toprule
Work & Principal target quantity & Information or target prior & Evidence relevant here\\
\midrule
Sharma et al.\ \citep{sharma2018} & Full relative pose & Known wireframe; monocular edges & Includes flight imagery\\
Capuano et al.\ \citep{capuano2020} & Shape and relative motion & Optical flow; star-tracker information & Numerical image simulations\\
Chi et al.\ \citep{chi2022} & Rotational and dynamic parameters & Monocular odometry; equal transverse inertia & Simulation\\
Zhang et al.\ \citep{zhang2023} & Pose and motion & Circular docking ring; nozzle; constrained dynamics & Simulation\\
Legrand et al.\ \citep{legrand2024} & Relative pose & Selected, pose-annotated target images; NeRF & SPEED+ HIL imagery\\
Kobayashi and Putkaradze \citep{kobayashi2026} & Normalized inertia & Attitude observations; torque-free arcs & Noise and image-derived attitude simulations\\
Francesch Huc and D'Amico \citep{huc2026} & Relative navigation & Learned odometry; known servicer state and maneuvers & Synthetic unseen targets\\
This study & Momentum-axis line and mean precession rate & Calibrated fixed camera; cylindrical geometry; principal-axis alignment & 96 synthetic scenes, two cold passes\\
\bottomrule
\end{tabularx}
\end{table}

\subsection{Rotational state, inertia and circular geometry}
The closest physical precedents do not all use spacecraft pose benchmarks. Chi et al.\ \citep{chi2022} filter monocular SLAM odometry and fit rotation vectors to a double-cone model for targets with equal transverse inertia. This directly precedes a claim about monocular recovery of symmetric-target rotational parameters. We have not implemented their estimator, and their parameter errors cannot be compared with our joint axis/rate threshold without matching the observations and definitions.

Hu et al.\ \citep{hu2024} examine rotational-parameter ambiguity in a stereovision estimator: 24 right-handed principal-axis frames can represent the same measurement history. They identify mass distribution using angular-momentum conservation and constrain the frame used by an EKF. This is a target-fixed coordinate ambiguity. Renaming a target basis preserves a physical momentum vector expressed in camera coordinates; reflecting that camera-space vector generally changes our requested quantity. Their observability argument does not establish identifiability of our endpoint fit.

Zhang et al.\ \citep{zhang2023} identify a docking ring as a spatial circle, use a nozzle feature to remove its pose duality and refine reprojection error before dynamical estimation. Their measurement has five degrees of freedom because rotation about the circle normal is unobserved, and an error-state filter includes inertia-ratio constraints. This is particularly close to our geometric motivation. Our current numerical loss emphasizes line endpoints; rim observations still enter geometry admission. We therefore claim neither the first use of circular spacecraft features nor the first combination of those features with tumbling dynamics.

Panicucci et al.\ \citep{panicucci2023} infer a small body's rotation axis from image feature tracks associated with circular landmark motion, selecting among geometric possibilities. Their asteroid problem differs in shape and visible texture, but it shares the need to distinguish a rotational quantity from a full pose. Circular-image inversion itself is ambiguous: Mariyanayagam et al.\ \citep{circle2018} discuss the twofold solution and calibration assumptions for a single circle. Temporal data may reduce a branch ambiguity, but only if the data and model contain discriminating information.

Inertial identification is another, stronger task. Pesce et al.\ \citep{pesce2017} use stereo measurements and filtering to estimate motion and inertia ratios. Candan and Servadio \citep{candan2026} combine monocular visual estimates with LiDAR in an augmented UKF. The extra depth channel changes the observability problem. Kobayashi and Putkaradze \citep{kobayashi2026} use attitude-only torque-free arcs, an initialization from constrained equations and nonlinear shooting with exact elliptic dynamics to estimate normalized inertia. Their image-derived attitudes are upstream measurements; our direct endpoint fit does not establish the same information content.

The torque-free equations and their analytic solutions are classical. Van Zon and Schofield \citep{vanzon2007} provide an exact rigid-body formulation and numerical implementation. Our use of Jacobi elliptic functions and real third-kind elliptic integrals draws on this established mathematics and the identities collected in DLMF \citep{dlmf}. What requires validation in our system is the connection from extracted image structure to the physical trajectory, not the existence of Euler--Poinsot solutions.

\subsection{Datasets and the simulation-to-reality gap}
SPEED+ explicitly tests transfer from synthetic training imagery to hardware-in-the-loop observations \citep{park2022speed}. SHIRT adds temporally structured rendezvous observations \citep{park2023ukf}. These resources provide stronger appearance-domain evidence than our all-synthetic study. However, neither can be inserted unchanged into our benchmark without defining an appropriate cylindrical target subset, motion reference and camera-motion handling.

Dataset names also conceal changing tasks. The original SPARK work \citep{musallam2021} concerns multi-modal spacecraft recognition, classification and detection. The later SPARK-2024 release provides pose-oriented sequences \citep{spark2024}. They should not be cited as interchangeable evaluations. SPE3R's multi-object shape emphasis \citep{park2024shape} and the 2026 SpaceSense-Bench preprint \citep{wu2026}, which supplies RGB, depth, LiDAR and semantic information for a larger object collection, address diversity beyond our four optical assemblies. We use these works to identify missing validation dimensions, not to imply that all their modalities or labels would be permissible inference inputs.

Our renderer varies geometry, viewing direction, lighting and material parameters, but includes no added sensor noise or motion blur. There are no flight images and no robotic-testbed measurements. A physically coherent renderer and conservation checks improve internal validity; they do not measure appearance transfer, sensor calibration error in operational conditions or the statistics of real debris.

\subsection{Selective prediction, uncertainty and evaluation integrity}
Selective prediction formalizes the trade-off between output coverage and error among accepted predictions \citep{el2010}. SelectiveNet \citep{geifman2019} learns prediction and rejection jointly. Our system instead uses a fixed conjunction of engineering gates. It is a selective estimator in the operational sense that it can withhold an answer, but it does not inherit the statistical guarantees of either framework.

Uncertainty estimation in spacecraft vision is an active topic. Dixon et al.\ \citep{dixon2025} use deep evidential regression to relate predicted uncertainty to pose error. Chen et al.\ \citep{chen2026} model pose uncertainty with unit dual-quaternion distributions and a multimodal formulation. Wang et al.\ \citep{wang2025} construct conformal keypoint sets and propagate sampled possibilities through PnP to pose sets. These approaches concern different uncertainty objects and assumptions. A retained bank of trajectories in our implementation is not automatically a calibrated prediction set.

Calibration asks whether reported probabilities correspond to empirical correctness, a distinction emphasized for learned models by Guo et al.\ \citep{guo2017}. Conformal risk control \citep{angelopoulos2024} provides a route to controlling expected loss under its calibration and exchangeability conditions. We have neither an untouched calibration population nor a validated loss family satisfying such a construction. Accordingly, our reports leave calibrated confidence and sigma fields null. A small local sensitivity or fold-to-fold radius is described as a diagnostic.

Finally, evaluation itself is an estimation procedure. Cawley and Talbot \citep{cawley2010} show how finite-sample model selection can bias subsequent performance assessment. Kapoor and Narayanan \citep{kapoor2023} document leakage in scientific uses of machine learning. Our pipeline is largely geometric, yet the same principle applies: selecting policies on opened cases and then treating those cases as confirmation is invalid. The study therefore separates development replays, a fresh population and repetitions of the same scenes. This separation is part of the result, not merely administrative bookkeeping.

\section{Problem, estimand and physical assumptions}\label{sec:problem}
Let $I_i$ denote a source image at acquisition time $t_i$, with known pinhole intrinsics $K$ and $t_0<\cdots<t_{F-1}$. The inputs include images, their acquisition calendar and calibration. Target attitude, inertia, momentum, stratum labels and reference motion do not enter inference. We write $R(t)\in\SO$ for body-to-camera orientation under a stationary camera and $a(t)=R(t)e_3$ for the longitudinal optical axis. The supported model assumes this optical axis coincides with a principal inertia axis.

For a torque-free rigid body, the body-frame angular velocity $\omega$ obeys
\begin{equation}
 I\dot{\omega}+\omega\times(I\omega)=0,\qquad
 \dot R=R[\omega]_\times .
 \label{eq:euler}
\end{equation}
The kinetic energy $E=\tfrac12\omega^\top I\omega$ and camera-frame angular momentum $H=RI\omega$ are constant. Define $h=H/\|H\|$. The requested axis is the line $[h]=\{h,-h\}\in\RP$, not a signed spin vector. Precession describes the azimuthal motion of $a(t)$ about $h$. With an orthonormal basis $u,v$ for $h^\perp$, a continuous azimuth is
\begin{equation}
 \phi(t)=\operatorname{unwrap}\operatorname{atan2}\bigl(v^\top a(t),u^\top a(t)\bigr),
 \qquad
 \bar{\Omega}=
 \left|\frac{\sum_i(t_i-\bar t)(\phi_i-\bar\phi)}
 {\sum_i(t_i-\bar t)^2}\right|.
 \label{eq:estimand}
\end{equation}
The implementation reports degrees per second. Here mean precession means the least-squares slope of unwrapped phase over the full source acquisition calendar, not an endpoint secant, instantaneous angular speed or slope only over frames admitted by feature extraction. For a nutating trajectory these definitions need not agree. When the transverse projection vanishes or the reference deems the rate unobservable, the positive-rate correctness criterion is unavailable. A nonzero visual frequency is not automatically this physical mean.

The inference contract is
\begin{equation}
 f(\{I_i,t_i\},K)\in
 \{(\widehat{[h]},\widehat{\bar{\Omega}}),\ \text{withheld},\
   \text{incomplete}\}.
\end{equation}
A withheld outcome is a completed decision that the declared checks do not support a point estimate. An incomplete outcome means execution or evidence could not establish that decision. Neither is counted as a correct estimate. Withholding also does not establish that the scene is physically unobservable: the extractor, search or gate policy may simply be inadequate.

We assume a rigid, approximately cylindrical target with visible geometric support; a fixed camera and target range; accurate acquisition times; supplied, undistorted calibration; and negligible external torque over the observation window. We do not infer metric target size or absolute inertia from the video. The triaxial parameterization permits unequal transverse inertia while preserving principal-axis alignment. Noncylindrical shape, optical/inertial misalignment and biased calibration are designated stress conditions, not silently incorporated into the supported domain.

\section{Measurement-to-decision method}\label{sec:method}
\subsection{Image measurements and their conditional nature}
The native C++20/OpenCV perception stage exports measured image geometry. Downstream code organizes raw line segments into alternative groups and measures candidate circular-rim geometry. Segments in one group are treated as simultaneous support, while groups within a frame represent competing interpretations. Collinear fragments are deduplicated using a fixed pixel-scale rule; their number is not treated as a count of independent temporal observations.

The strongest unresolved assumption is feature identity. A mathematically correct line projection formula cannot determine whether an extracted line is a physical generator. Similarly, a fitted ellipse need not be the projection of a circular rim. Geometry admission recomputes selected training decisions, but the preceding feature extraction consumes the whole video, including stateful glare preprocessing. Thus, leave-block-out checks later in the pipeline are conditional on shared measured features. They are not pixel-isolated end-to-end cross-validation.

\begin{figure}[t]
\centering
\begin{tikzpicture}[node distance=5mm,box/.style={draw=teal!65!black,rounded corners,align=center,text width=125mm,minimum height=9mm,font=\small},arr/.style={-{Latex[length=2mm]},thick,draw=teal!65!black}]
\node[box] (input) {Images, source timestamps and camera calibration\\No reference poses, inertias or stratum labels};
\node[box,below=of input] (features) {Whole-video image measurements: lines, alternative groups and rims};
\node[box,below=of features] (search) {TRAIN geometry admission and complete finite banks\\480 cone starts + 128 triaxial starts; fixed family selection};
\node[box,below=of search] (full) {Full-fit numerical and geometric checks\\Failed check: completed withholding; missing evidence: incomplete};
\node[box,below=of full] (folds) {Only after full-fit checks pass: eight fresh temporal fits\\Shared upstream features; selected fit and retained rivals assessed};
\node[box,below=of folds] (out) {Linked axis/rate estimate if all 27 checks pass\\Otherwise withholding or explicitly incomplete execution};
\foreach \a/\b in {input/features,features/search,search/full,full/folds,folds/out}\draw[arr] (\a)--(\b);
\end{tikzpicture}
\caption{The evaluated decision flow. Fresh temporal fits reset numerical search; they do not remeasure isolated pixel folds. The diagram describes computation and evidence, not a guarantee that admitted features have the correct physical identity.}\label{fig:pipeline}
\end{figure}
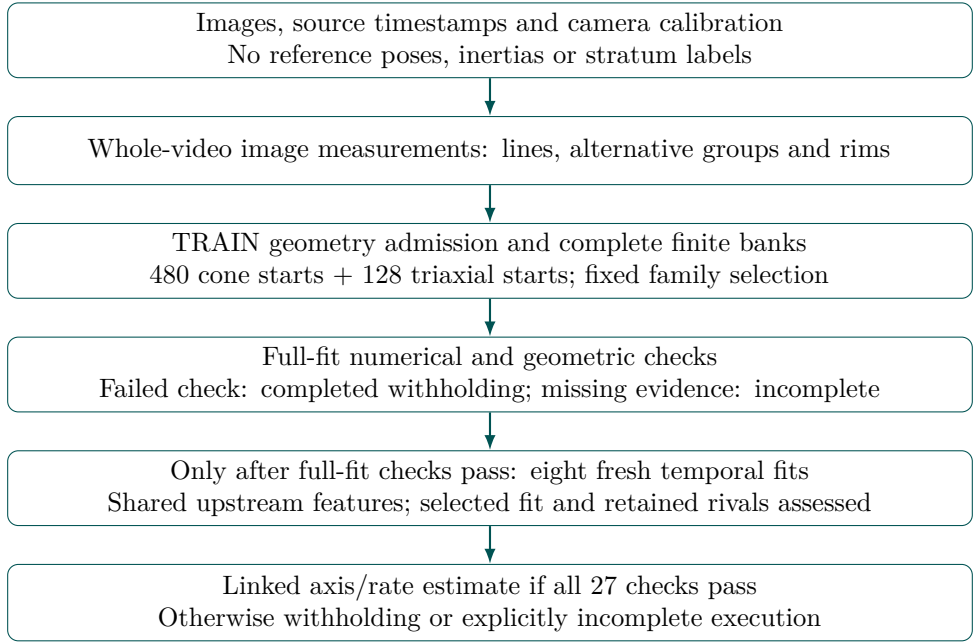

\subsection{Endpoint projection and positive-ray support}
A 3D direction $a(t)$ has homogeneous vanishing point $q(t)=Ka(t)$. For an observed segment with inhomogeneous endpoints $p_1,p_2\in\R^2$, let $\ell=\widetilde p_1\times\widetilde p_2$ and $d_j=q_{xy}-q_zp_j$. Define $S=d_1d_1^\top+d_2d_2^\top$. The free-line endpoint correction is
\begin{equation}
 d_{\rm line}^2(q,p_1,p_2)=
 \frac{(\ell^\top q)^2}{\lambda_{\max}(S)}.
 \label{eq:line}
\end{equation}
This is a pixel-space correction, not an angular uncertainty. It has a finite parallel-line limit when $q_z=0$. For $q_z\ne0$, physical generator endpoints should lie on a common ray from the vanishing point under the declared geometry. The implemented profile is
\begin{equation}
 d_{\rm ray}^2=
 \begin{cases}
 d_{\rm line}^2,& d_1^\top d_2>0,\\
 \min(\|d_1\|^2,\|d_2\|^2)/q_z^2,&d_1^\top d_2\leq0 .
 \end{cases}
 \label{eq:ray}
\end{equation}
At $q_z=0$, the first branch applies. In the second branch an endpoint reaches the vanishing point; this is generally an infimum for distinct endpoints at finite positive depth. Separate support gates therefore matter. Appendix~\ref{app:geometry} derives these formulas and states their limits. The profile is sign-invariant in $q$, consistent with an unoriented longitudinal direction.

For frame $i$, candidate group $g$ contains $m_{ig}$ deduplicated segments. With $\sigma_{\rm px}=1$ pixel and Huber transition $\delta=2.5$, the working objective is
\begin{align}
 \rho_\delta(r)&=
 \begin{cases}r^2,&|r|\leq\delta,\\2\delta|r|-\delta^2,&|r|>\delta,\end{cases}\\
 L(\theta)&=\frac1N\sum_{i\in\mathcal T_{\rm line}}
 \min_{g\in\mathcal G_i}\frac1{m_{ig}}
 \sum_{j=1}^{m_{ig}}
 \rho_\delta\!\left(\frac{d_{\rm ray}(Ka_\theta(t_i),p_{igj,1},p_{igj,2})}
 {\sigma_{\rm px}}\right).
 \label{eq:loss}
\end{align}
Here $N$ is the number of distinct admitted training line frames. Averaging within a group prevents raw fragment count from directly multiplying frame weight; the minimum across groups implements an OR over interpretations. It also permits a wrong but coherent group to dominate. The pixel scale is fixed, not estimated or calibrated from sensor measurements. The robust minimum-group objective is not a demonstrated likelihood.

In the evaluated candidate, rim-normal residuals are excluded from both training and held-out motion losses. Rims nevertheless participate in training geometry admission. Calling this a rim-free estimator would therefore be incorrect. Both motion families use the same canonical line objective after the implementation's recorded cost normalization.

\subsection{Constant-cone and torque-free triaxial families}
The constant-cone family is
\begin{equation}
 a(t)=h\cos\alpha+
       \sin\alpha\bigl(u\cos(\Omega t+\phi_0)+v\sin(\Omega t+\phi_0)\bigr).
 \label{eq:cone}
\end{equation}
The two axis degrees of freedom, cone aperture, frequency and phase give five nominal parameters. Unlike an arbitrary affine image-direction ellipse, this construction enforces a physical circular cone in 3D. The complete bank contains 480 initialized fits. The search is finite and locally optimized; complete bank means all prescribed starts were retained, not that all physical solutions were found.

The triaxial family normalizes $I_3=1$ and sets $I_1=J+d/2$, $I_2=J-d/2$, with $I_1\geq I_2>1$ and $0<d<1$. The difference bound enforces the relevant strict triangle inequality. Let
\begin{equation}
 A=1/I_1,\quad B=1/I_2,\quad \varepsilon=2E/\|H\|^2,\quad
 b^2=\frac{\varepsilon-A}{1-A},\quad
 \gamma=\frac{\varepsilon-B}{1-B}.
\end{equation}
An interior energy fraction selects $\varepsilon\in(A,B)$ for the crossing (CN) regime or $\varepsilon\in(B,1)$ for the noncrossing (DN) regime. In a momentum-aligned frame, the longitudinal coordinate takes the form
\begin{equation}
 z(t)=
 \begin{cases}
 b\,\operatorname{cn}(\nu t+\eta_0\mid m),&
 m=b^2/(b^2-\gamma),\quad\text{CN},\\
 s\,b\,\operatorname{dn}(\nu t+\eta_0\mid m),&
 m=1-\gamma/b^2,\quad s\in\{-1,+1\},\quad\text{DN}.
 \end{cases}
\end{equation}
The signed auxiliary quantity $\gamma$ is negative in CN and positive in DN. The scale is $\nu=\|H\|\sqrt{(1-B)(1-A)(b^2-\gamma)}$ for CN and
$\nu=\|H\|\sqrt{(1-B)(1-A)b^2}$ for DN. The azimuth satisfies
\begin{equation}
 \dot\phi=\|H\|\frac{\varepsilon-z^2}{1-z^2},\qquad
 a_{\rm local}(t)=
 \left(\sqrt{1-z^2}\cos\phi,\sqrt{1-z^2}\sin\phi,z\right)^\top .
 \label{eq:triax}
\end{equation}
A fixed three-parameter rotation maps this path to camera coordinates. Together with energy fraction, positive scale parameter, $J$, $d$ and phase fraction, there are eight nominal continuous parameters. The search scale is not itself the reported mean precession rate; the latter is computed from the linked path over the full source span.

The real third-kind integral implements the azimuth using Carlson reduction and explicit period counting \citep{dlmf}. Independent quadrature refinement is checked for the selected triaxial path. The 128 prescribed starts include the two regimes and their specified signs. All outcomes, including optimizer nonconvergence, are retained. These equations encode classical torque-free motion \citep{vanzon2007}; they do not prove that image observations identify every parameter or exclude trajectories outside the bank.

The search also has fixed parameter bounds and numerical budgets. In particular, its angular-scale ceiling is 30 degrees/s; the triaxial scale parameter is not the reported mean rate. Appendix~\ref{app:numerics} specifies the bounds, initializations, solver stages and termination rules. These restrictions are part of the evaluated estimator, rather than a claim to search all torque-free trajectories.

\subsection{Fixed family selection and separate acceptance}
For each family $k$, the minimum raw-cost fit is selected from its complete bank, including nonconverged fits. The family score is
\begin{equation}
 S_k=N\log\!\bigl(\max(L_k,10^{-12})\bigr)+p_k\log N,\qquad
 p_{\rm cone}=5,\quad p_{\rm triaxial}=8.
 \label{eq:selection}
\end{equation}
The canonical factors recorded in the implementation are one for cone and two for the raw triaxial cost. Ordinal ties within a bank retain the first fit; an exact family-score tie selects cone. The form is motivated by dimension-penalized selection \citep{schwarz1978}, but the statistical derivation of a Bayesian information criterion does not apply automatically to correlated frames, a robust OR loss or nominal dimensions near degeneracy. We use it as a frozen engineering rule.

Acceptance cannot alter the selection. If the chosen minimum fails convergence or a required support check, a more convenient runner-up is not substituted. This prevents acceptance-aware fallback from silently changing the estimator being evaluated. It can also lower coverage, especially if numerical convergence diagnostics are conservative or the local search behaves poorly. The study does not isolate that effect causally.

\subsection{Temporal checks and competing trajectories}
The full-data fit first undergoes phase, sampling, geometry, convergence and numerical screening. A failed full-fit gate is sufficient for a completed withholding decision; eight temporal blocks may then remain explicitly unexecuted. If the full checks pass, the source calendar is split into eight contiguous blocks. For each block, a fresh complete search fits the complementary training frames and evaluates the held-out block. Selected parameters, training quality, held-out support and retained rivals are all recorded.

Two points are considered materially distinct when their sign-only axis distance is at least $10^\circ$ or their symmetric rate difference,
$200|\Omega_1-\Omega_2|/(\Omega_1+\Omega_2)$, is at least $15\%$. This engineering comparison of candidates differs from reference-relative rate error used in evaluation. Training-compatible distinct rivals must be predictively worse under the frozen rules; unresolved rival optimization can veto an estimate.

All 27 declared gates must be supported before a point is emitted. Their inventory appears in Appendix~\ref{app:gates}. They include the six phase/sampling checks, full and nested convergence, pixel and angular support, positive-ray support, numerical conditioning screens, eight completed blocks, nonempty held-out observations, conditional parameter stability and rival checks. The resulting radii summarize sensitivity within these calculations. They are neither credible intervals nor confidence regions. Appendix~\ref{app:decisiondiagnostics} gives their exact dispersion construction and the rival numerical rules. Missing or malformed required evidence produces an incomplete decision rather than an invented gate value.

\section{Evaluation design and provenance}\label{sec:design}
\subsection{New synthetic population}
The final study fixed seed 2026102501 before generation and registered 96 scenes: 32 axisymmetric core, 16 triaxial CN, 16 triaxial DN and 32 stress scenes. The supported subset is the first 64 scenes. Stress consists of four scenes for each of eight prespecified interventions. Four optical assemblies recur across strata; the scenes are new draws within this small geometry bank, not 96 independently designed target shapes.

Each scene contains 180 images at $768\times768$ pixels. Acquisition frequency is drawn independently of motion from a uniform distribution on $[0.9,1.1]$ frames/s, using a domain-separated pseudorandom stream. Consequently, source spans are approximately 163--199 seconds. The nominal rotational scale is sampled from $[5,24]$ degrees/s; it is a generation parameter, not necessarily the evaluated phase-regression mean. Momentum-axis view angles are drawn from $25^\circ$--$70^\circ$. The fixed range is sampled from 42--90 m and focal length from 42--62 mm.

The generator uses Blender/Cycles with 32 rendering samples, varying material parameters, solar geometry and exposure. Supported optical assemblies are built from positive-mass primitives with internal sphere pairs used to realize inertia differences. The four body diameter/length pairs are 3.2/8.1, 4.4/10.8, 3.8/9.8 and 3.6/10.6 m; appendage and rim dimensions vary as well. This is an idealized mass construction, not a measured spacecraft inertia model. There is no added sensor noise or motion blur.

Reference dynamics use a separate split-Hamiltonian rigid-body integrator rather than importing the inference Jacobi path. The generator specifies 128 integration substeps per acquisition interval, doubled for the near-separatrix strata, and eightfold temporal oversampling for diagnostics. Its conservation admission thresholds are $10^{-5}$ relative energy drift and $10^{-8}$ relative momentum drift. Source identity and render/runtime bindings are archived. This separation reduces one form of implementation circularity, while leaving common physical assumptions and renderer limitations intact.

Boundary CN/DN cases approach the separatrix; the normalized separation draw changes from $[0.15,0.85]$ to $[0.01,0.05]$. Low-light cases alter illumination and exposure; sparse-rim cases reduce rim geometry. Outside-domain cases introduce biased supplied intrinsics, coherent distractors, noncylindrical deformation or optical/principal-axis misalignment. For the intrinsics intervention, supplied $f_x$ is multiplied by 1.08, $f_y$ by 0.94, and the principal point is shifted by $(12,-9)$ pixels. These are specific interventions, not a representative sample of all possible deployment shifts.

\subsection{Isolation, compute and repetitions}
Generation, inference and scoring are separate stages. Inference receives opaque scene identifiers, pixels and allowed acquisition metadata; reference states and the scorer are not mounted. Each attempt uses a fresh pinned container with four CPU units, 12 GiB memory, a 7200 s limit, no network, a read-only root and source/input mounts, and only its own output writable. At most four outer attempts run concurrently. Up to four fresh inner fold processes are permitted, with numerical-library threads fixed to one.

The first pass is primary. All first-pass attempts finish before the second begins. Both passes repeat the complete image-to-report computation, rather than only rerunning a score on cached estimates. The 192 attempts are therefore 96 scene pairs, not 192 independent samples of physical accuracy. Missing or failed execution, preparation and evidence remain in the fixed denominator. No case is replaced, retried for a better score or reassigned to a more favorable stratum.

The protocol was internally frozen and committed before rendering, including sources, resource limits, the scientific comparator and the attempted-case inventory. This is an internal prospective registration, not an external registered report or independent institutional audit. The same project developed the generator and estimator. Source separation prevents direct reference input to inference; it does not establish independence of all scientific design choices.

\subsection{Linked joint metric and outcome accounting}
For unit reference and estimated axes, the metric is
\begin{equation}
 e_h=\frac{180}{\pi}\arccos\!\left(\operatorname{clip}(|h^\top\hat h|,0,1)\right),
 \qquad
 e_\Omega=100\,\frac{|\widehat{\bar\Omega}-\bar\Omega|}{\bar\Omega}.
 \label{eq:metric}
\end{equation}
The rate metric is used only for an observable positive reference rate. Joint correctness requires $e_h<10^\circ$ and $e_\Omega<15\%$ on the \emph{same linked emitted estimate}. An accurate rate from one hypothesis and an accurate axis from another do not form a success. A global depth reflection is not an allowed equivalence.

For $n$ planned scenes, let $C,W,A,F$ count correct estimates, wrong estimates, valid withholding decisions and incomplete executions. The accounting identity is
\begin{equation}
 C+W+A+F=n,\qquad
 \widehat{\mathrm{coverage}}=(C+W)/n,\qquad
 \widehat{\mathrm{correct\ coverage}}=C/n.
\end{equation}
The empirical error conditional on emission is $W/(C+W)$ when the denominator is nonzero. A method can achieve zero observed conditional error by refusing nearly everything; coverage and failure counts are therefore essential descriptive context.

The registered targets were at least 20/32 correct core estimates, at least 16/32 combined triaxial estimates, at least 6/16 in each triaxial regime, zero wrong supported emissions, zero execution/evidence failures over all 192 attempts and established scientific repetition for every scene. Stress cases remain reported but do not substitute for missing supported successes.

\subsection{Original failure and descriptive supplement}
The original observer chain did not close as a successful validation. A later native, authenticated descriptive scoring supplement accounts for all attempts and produces the counts reported here. It preserves the original completion evidence and failures; it does not retroactively satisfy the registered execution or repeatability targets. The supplement's score digest is recorded in Appendix~\ref{app:provenance}.

The protocol also planned a genuine-v4 comparison on the same 96 scenes. Preparation and watcher-launch evidence exist, but a completed, authenticated same-population baseline score is not present in the closed evidence used by this manuscript. That planned comparison is therefore not reported as a completed experiment. Historical v4 scores belong to a different population. This omission limits causal and comparative conclusions; the paper is a candidate case study rather than a completed two-arm benchmark.

\section{Results}\label{sec:results}
\subsection{Primary-pass accuracy, withholding and incomplete execution}
\begin{figure}[ht]
\centering\includegraphics[width=.96\linewidth]{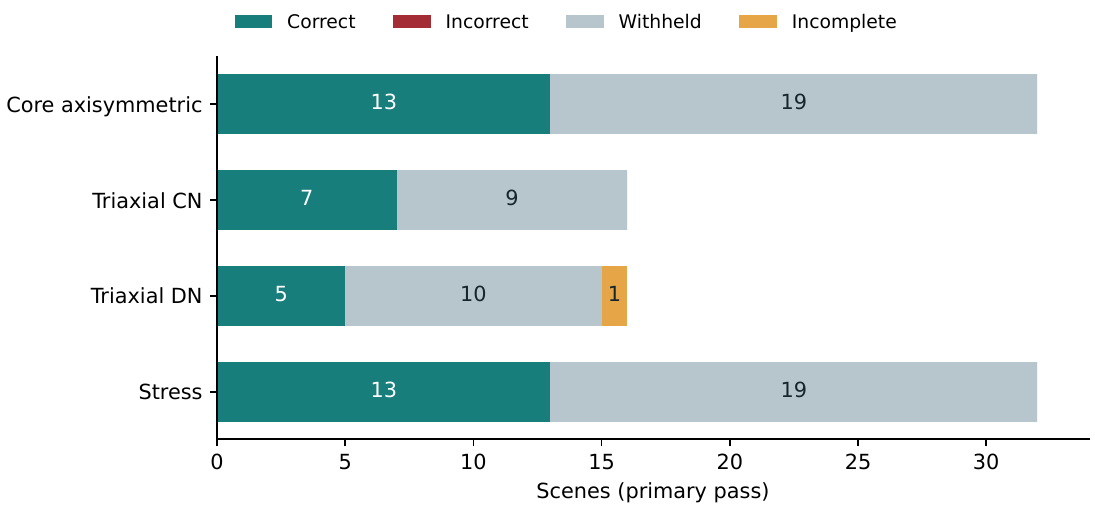}
\caption{Outcome counts by the three supported strata and the stress subtotal. The accepted subset is small despite zero observed wrong emissions.}\label{fig:outcomes}
\end{figure}

Table~\ref{tab:primary} and Figure~\ref{fig:outcomes} report every scene in the primary pass. There are 38 correct estimates, zero wrong estimates, 57 valid withholding decisions and one incomplete execution. Correct coverage is 39.6\% overall and 39.1\% on the supported subset. Since no incorrect point was emitted, emission coverage and correct coverage coincide in this sample. This equality is an empirical result, not a property guaranteed by the gates.

\begin{table}[ht]
\centering\small
\caption{Primary pass on the new 25 October population. Incomplete execution is a separate outcome, included in every applicable denominator. Coverage is correct joint estimates divided by planned scenes.}\label{tab:primary}
\begin{tabular}{lrrrrrr}
\toprule
Population & $n$ & Correct & Wrong & Withheld & Incomplete & Cov. (\%)\\
\midrule
Core axisymmetric & 32 & 13 & 0 & 19 & 0 & 40.6\\
Triaxial CN & 16 & 7 & 0 & 9 & 0 & 43.8\\
Triaxial DN & 16 & 5 & 0 & 10 & 1 & 31.2\\
Supported subtotal & 64 & 25 & 0 & 38 & 1 & 39.1\\
Stress subtotal & 32 & 13 & 0 & 19 & 0 & 40.6\\
All scenes & 96 & 38 & 0 & 57 & 1 & 39.6\\
\bottomrule
\end{tabular}

\end{table}

Only the CN coverage check and the zero-observed-wrong-supported check pass. Core coverage is seven successes below its target; combined triaxial coverage is four below; DN coverage is one below. The two failed attempts violate the failure-free target. One incomplete scene pair prevents the all-scene scientific-repetition target from passing. The registered conjunction therefore fails.

\subsection{Errors among emitted estimates}
Post-closure descriptive analysis of the closed score gives a median sign-only axis error of $0.240^\circ$ and a maximum of $1.997^\circ$ among the 38 emissions. Median rate error is $0.00169\%$, with maximum $0.03347\%$. The percentage convention was checked against the frozen metric adapter: the stored rate field includes multiplication by 100. These are selected-subset statistics under the synthetic generator, not errors over all scenes.

\begin{table}[ht]
\centering\small
\caption{Errors conditional on emission, primary pass only. Rates are percentage errors, not fractions. These descriptive summaries were computed after study closure; they are not additional registered endpoints.}\label{tab:errors}
\begin{tabular}{lrrrrr}
\toprule
Population & Emitted & $e_h$ median ($^\circ$) & $e_h$ max ($^\circ$) & $e_\Omega$ median (\%) & $e_\Omega$ max (\%)\\
\midrule
Core axisymmetric & 13 & 0.302 & 1.997 & 0.00112 & 0.00865\\
Triaxial CN & 7 & 0.254 & 1.287 & 0.00207 & 0.03347\\
Triaxial DN & 5 & 0.547 & 1.246 & 0.00255 & 0.00620\\
Stress & 13 & 0.171 & 0.599 & 0.00279 & 0.02018\\
All & 38 & 0.240 & 1.997 & 0.00169 & 0.03347\\
\bottomrule
\end{tabular}

\end{table}
\begin{figure}[ht]
\centering\includegraphics[width=.83\linewidth]{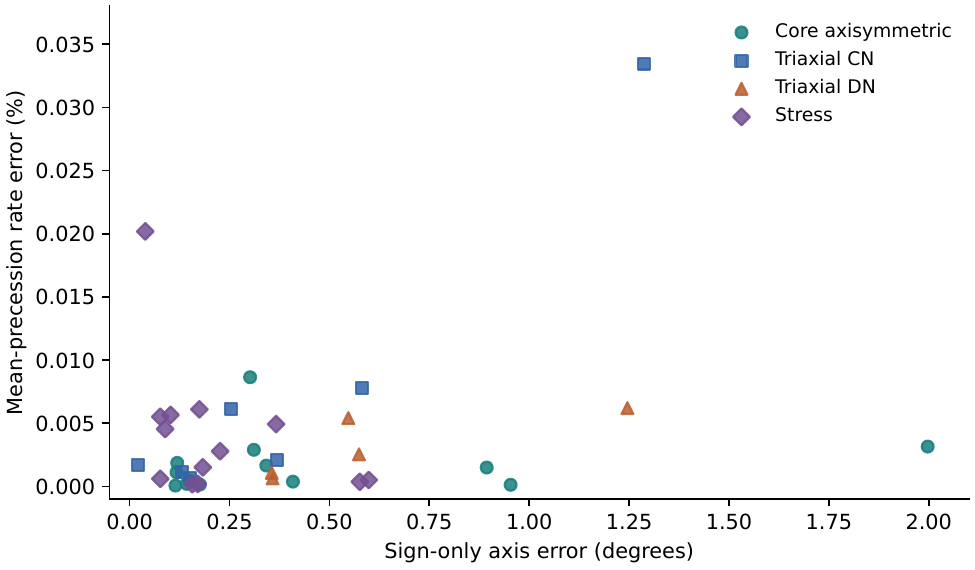}
\caption{The 38 emitted estimates. Axes are deliberately zoomed to show the observed errors; correctness limits are $10^\circ$ and 15\%, well outside this view. The 57 withheld scenes and the incomplete scene have no plotted point estimate.}\label{fig:errors}
\end{figure}

The small errors deserve neither dismissal nor overinterpretation. Long synthetic sequences without added sensor noise can support precise fits when the selected features agree with the model. At the same time, acceptance selects favorable cases. We cannot assign the accepted-subset error distribution to the withheld population, infer sensor-level calibration from it or compare it directly with per-frame attitude errors in the literature.

\subsection{Stress outcomes}
Table~\ref{tab:stress} keeps all eight interventions visible. All four low-light, all four biased-intrinsics and all four optical-axis-misalignment scenes are withheld. Some separatrix-boundary, sparse-rim, distractor and noncylindrical cases yield correct estimates. Four scenes per intervention are too few to establish a general rejection detector or a robust out-of-domain accuracy rate.

\begin{table}[ht]
\centering\small
\caption{Prespecified stress scenes. Correct emissions outside the supported model do not expand that model's validated domain.}\label{tab:stress}
\begin{tabular}{lrrrrr}
\toprule
Intervention & $n$ & Correct & Wrong & Withheld & Incomplete\\
\midrule
Near-boundary CN & 4 & 3 & 0 & 1 & 0\\
Near-boundary DN & 4 & 4 & 0 & 0 & 0\\
Low light & 4 & 0 & 0 & 4 & 0\\
Sparse rims & 4 & 3 & 0 & 1 & 0\\
Biased intrinsics & 4 & 0 & 0 & 4 & 0\\
Coherent distractor & 4 & 1 & 0 & 3 & 0\\
Noncylindrical & 4 & 2 & 0 & 2 & 0\\
Optical-axis misalignment & 4 & 0 & 0 & 4 & 0\\
\bottomrule
\end{tabular}

\end{table}

The stress correct-coverage rate is slightly above the supported rate. This does not imply that the stress population is intrinsically easier or that model violations improve estimation. These are small, fixed strata with different nuisance realizations. Neither a matched intervention contrast nor a significance claim is supported by the aggregate table.

\subsection{Repetition and the incomplete scene}
The second pass has the same aggregate outcome counts. The frozen scientific comparison establishes equivalence for 95 pairs, establishes no differing pair and leaves one pair non-comparable because both attempts were incomplete. This is not 96/96 reproducibility, and the absence of an established difference does not convert the non-comparable pair into a match.

The incomplete scene, \texttt{pv2-54f18b60cc016565}, belongs to the supported DN stratum. Seven temporal blocks completed. Preparation of zero-indexed block 7 failed with \texttt{Insufficient\-Pixel\-Data\-Error}. Both attempts exited with code 2, without timeout, and cleanup was authenticated. The runtime did not establish a valid terminal withholding decision. We therefore retain the attempts as execution failures. A later admission-contract repair would be a new method version, not grounds to rewrite this result.

Scientific equality is defined by the frozen comparator, not by a retrospective numerical tolerance. The policy permits narrow, typed operational differences such as process identifiers and times while retaining physical parameters, costs, optimizer flags, calendars, decisions, rivals and unknown structured fields. It also excludes three authenticated perception-stage standard-error logs, corresponding to mask, spatial and ellipse extraction, from the scientific view. The raw logs remain retained and binding for raw equality; the exclusion does not apply generally to numerical child outputs. Raw artifact equality is therefore a different and stricter notion. The exact comparator and policy accompany the source package. Repetition on one pinned environment also falls short of reproduction by an external research group.

\subsection{What zero observed errors can support}
For intuition only, under an idealized iid Bernoulli model with a fixed accepted-population error probability $p$, zero errors in $m$ emissions yields the exact one-sided 95\% upper bound
\begin{equation}
 p_{\rm upper}=1-0.05^{1/m}.
\end{equation}
This is approximately 7.58\% for $m=38$ and 11.29\% for the 25 supported emissions. Those values are not calibrated uncertainty attached to an individual estimate. More importantly, the actual study has fixed strata, repeated geometries, shared rendering assumptions and selection by a complex gate. We present the calculation as an illustration of finite-sample limitations, not as a certified pooled population or flight-risk bound. Counting the repeated pass as 76 independent emissions would be unjustified.

\section{Historical corrections and negative controls}\label{sec:corrections}
\subsection{An axis metric can erase a physical error}
\begin{figure}[ht]
\centering\includegraphics[width=.82\linewidth]{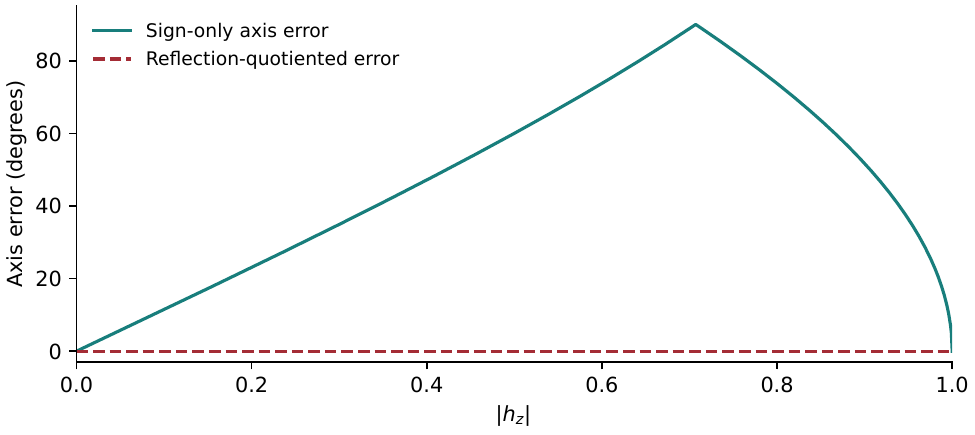}
\caption{Analytical diagnostic, not a new empirical experiment. A depth-reflected axis can have zero error under an overly permissive quotient while being orthogonal as an unoriented camera-frame axis.}\label{fig:reflection}
\end{figure}

An earlier development score permitted both global sign and a depth reflection. Under sign-only scoring, the reported 13/13 development result becomes 11/13. The difference is not cosmetic. Let $M=\operatorname{diag}(1,1,-1)$ and suppose the estimator returns $\hat h=Mh$. A reflection-quotiented metric can assign zero error, whereas the sign-only metric is
\begin{equation}
 d_{\RP}(h,Mh)=\arccos|h^\top Mh|
             =\arccos|1-2h_z^2|.
 \label{eq:reflection}
\end{equation}
It reaches $90^\circ$ when $h_z^2=1/2$. Figure~\ref{fig:reflection} illustrates this elementary counterexample. The correct equivalence must follow the intended physical quantity; it cannot be enlarged merely because an estimator confuses branches.

\subsection{Acquisition cadence and opened development data}
The earlier 48-video generator selected acquisition cadence using true rotational motion. Its metadata therefore exposed a rate proxy. Correcting the axis convention changed a historical 38/48 score to 25/48, but metric correction alone cannot remove the acquisition confound. Those results are not independent evidence that images identify the rate. The final generator separates cadence from motion and fixes the frame count before inference.

After opening the 24 October population, a successor policy produced 51 correct candidate readouts in an archived-bank development replay. Because those cases had already informed the policy, 51/96 is development evidence. It is not the result of the new study, not an estimate of generalization and not a substitute for the final 38/96. This distinction applies even if every numerical fit and reference is correctly linked.

\begin{table}[ht]
\centering\small
\caption{Evidence populations must not be merged. Historical rows describe their original studies or corrected development scores; they are not additional independent samples for the final candidate.}\label{tab:history}
\begin{tabularx}{\linewidth}{>{\raggedright\arraybackslash}p{29mm}>{\raggedright\arraybackslash}p{36mm}Y}
\toprule
Population & Recorded result & Interpretation\\
\midrule
13 development cases & 13/13 becomes 11/13 under sign-only axis scoring & Old reflection equivalence inflated success; reused development data\\
48 earlier videos & 38/48 becomes 25/48 under strict scoring & Motion-dependent cadence remains a confound\\
24 Oct, 96 scenes, v4 & 38 correct, 36 wrong, 22 withheld & Predecessor on a different population\\
24 Oct, physical endpoint & 15 correct, 0 wrong, 63 withheld, 18 timeouts & Same historical study; registered targets failed\\
Opened 24 Oct bank replay & 51 correct, 0 wrong, 45 withheld & Policy development after outcomes opened\\
25 Oct, 96 new scenes & 38 correct, 0 wrong, 57 withheld, 1 incomplete & Present frozen-candidate descriptive result\\
\bottomrule
\end{tabularx}
\end{table}

\subsection{Feature identity, closure and aliasing}
Historical image-direction probes found coherent bias that left a high fit score despite a wrong axis. Under an in-plane rotation applied consistently to every measured direction, an equivariant trajectory fit can rotate its recovered axis while preserving its residual structure. Temporal prediction from those same directions then reproduces the bias. This is a measurement-identification problem, not simply an optimizer robustness problem.

Specific extractor controls made the problem concrete: internal edge populations could imitate a longitudinal direction, and a long-segment rule did not eliminate coherent long texture. An earlier affine harmonic fit also admitted a direction ellipse whose offset was incompatible with a physical cone. Circular harmonic arms must be orthogonal and equal in 3D, and their offset must align with their normal. Satisfying only the arm constraints is insufficient. The current physical-family construction addresses that particular model inconsistency, while leaving feature identity unresolved.

Regular sampling introduces another ambiguity. A historical projected-direction example used a true rate of $25.90586^\circ$/s with 20 s sample spacing and produced $7.90586^\circ$/s with a high diagnostic score. The later regular-lattice guard checks retained timestamps and conservatively excludes an unsigned search range extending beyond $90/\Delta t$ degrees/s. A tolerant lattice-gap calculation handles missing ticks. This engineering guard addresses the demonstrated alias; it is not a theorem of uniqueness for irregular samples or all physical trajectories.

These controls were conducted on historical development inputs or analytical observations. They explain failure mechanisms, but they do not add cases to the final prospective denominator or establish how often those mechanisms occur in flight imagery.

\subsection{Numerical improvement is not physical improvement}
A separate paired control restarted 36 already converged checkpoints: four full fits and 32 training fits from four development sequences. Scalar and vector residual formulations produced 72 completed endpoints. Vector residuals reduced canonical SSE in every paired comparison, but the median relative decrease from the original checkpoint was only $7.85\times10^{-9}$ and the maximum was $2.88\times10^{-7}$. Objective callbacks increased by 30.7\%. Maximum axis displacement was $0.001102^\circ$.

Those are local numerical effects on correlated checkpoints. They rescue no failed original winner, establish no reduction in error against truth and supply no new coverage result. The evaluated candidate was not changed on that basis. Preserving this distinction avoids using an optimizer's success flag or a tiny objective decrease as a surrogate for scientific progress.

\section{Discussion}\label{sec:discussion}
\subsection{What is learned}
The final experiment demonstrates that the implemented measurement and decision chain can produce precise, repeatable joint estimates for a subset of new synthetic scenes while withholding many others. It also shows that the chosen policy fails its own minimum-coverage and execution requirements. Both observations are needed: reporting only emitted precision would overstate usefulness, while reporting only the failed target would conceal the functioning subset.

The contribution with the clearest support is methodological and empirical. The same project records a progression from permissive evaluation and metadata confounding to a stricter, failure-preserving experiment. The endpoint objective, linked estimate identity, fixed family selector and explicit incomplete state make that experiment inspectable. We do not establish that this combination is the first of its kind, that every gate is necessary or that a simpler baseline would perform worse.

The gap between low accepted error and low coverage suggests several possible bottlenecks: missing valid image structure, overly restrictive admission, optimizer nonconvergence, unresolved competing fits and conservative temporal checks. The current aggregate score does not separate their causal contributions. Relaxing gates after seeing the new cases might improve a replay count but would reopen development; another untouched population would be needed for confirmation.

\subsection{Four distinct notions of confidence}
Physical correctness, numerical convergence, temporal stability and execution reproducibility answer different questions. Convergence concerns a numerical stopping rule. Temporal stability concerns sensitivity to a particular partition and measurement representation. Reproducibility concerns whether the same declared computation yields equivalent scientific outputs. Physical correctness concerns the target quantity relative to a valid reference. None of the first three alone implies the fourth.

The same separation applies to geometry. A positive-ray fit is more constrained than an arbitrary image line, but still presumes that the line belongs to the object structure the model describes. A finite list of surviving alternatives is useful evidence about the search that was performed. It does not certify that no unsearched alternative, different shape or illumination-dependent edge could explain the images.

\subsection{Threats to validity}
\paragraph{Internal validity.}
The reference dynamics are implemented separately from inference, and labels are absent from inference mounts. Nevertheless, both components share the torque-free modeling premise and project-developed geometry conventions. Software tests and source hashes can bind an implementation; they cannot independently validate the physical meaning of every feature or reference.

\paragraph{Selection and preprocessing.}
The final population was generated after the candidate freeze, but the method was developed extensively on earlier cases. The 27-gate conjunction embodies those choices. Shared whole-video preprocessing also means temporal checks cannot be interpreted as independent held-out pixel tests. Scene-level prospective evaluation is the relevant protection against directly reusing the final outcomes; it does not eliminate shared model misspecification.

\paragraph{Statistical precision.}
There are only 96 scene draws from four assemblies and only four cases per stress intervention. The repeated pass is not another population. No calibrated probability model for accepted error has been fitted or validated. The small conditional-error summaries should not be extrapolated to a deployment distribution.

\paragraph{External validity.}
There are no real sensor images, motion blur, injected sensor noise, flexible appendages, uncontrolled camera motion, sustained external torques or general noncylindrical targets in the supported experiment. Unknown real inertia and material properties are replaced by an idealized construction. No capture, collision-avoidance or onboard real-time claim follows.

\paragraph{Comparative validity.}
The planned same-population predecessor comparison has no completed score in the closed evidence used here, and no external estimator was run. Different historical populations cannot identify the effect of the new policy. Literature methods often estimate full pose under different sensing or target assumptions; their published errors are not interchangeable with our metric.

\paragraph{Reproduction and availability.}
The scientific repetition used a pinned environment within the project. It was not performed by an independent laboratory. The source package includes the original closed score in losslessly compressed form, the protocol, the attempted-case registration and frozen scientific source excerpts. At this manuscript version, there is no public archival DOI for the complete image-to-report evidence. The supplied score and figure-generation source support checking the descriptive arithmetic. Archived reports and reference trajectories also permit recomputation of the physical errors. Repeating image-to-report inference still requires the omitted images, runtime and remaining dependencies.

\subsection{Experiments that would materially strengthen the claim}
A credible next evaluation would first execute a matched baseline on a new or still-unopened population with an explicit common estimand. Useful arms include a constant-cone-only policy, the fixed triaxial family, the complete selector and an adapted published rotational-state method. Each arm should preserve the same pixels, calibration, resource budget and incomplete-attempt accounting. Post-hoc oracle family selection should be reported only as an oracle diagnostic.

Feature identity requires a separate intervention. Annotated generator, rim and texture correspondences could distinguish absence of a correct candidate from a selector choosing the wrong candidate. Matched renders with controlled distractors, followed by laboratory imagery of a rotating cylindrical mockup, would test whether temporal consistency survives realistic specular changes. Pixel-fold-local preprocessing would be needed before claiming end-to-end temporal holdout.

Uncertainty work should use a separate calibration split and an untouched test split, state the population and exchangeability assumptions, and report risk versus coverage under a frozen selection rule. A heuristic gate score must not be renamed a probability. For a sequential navigation application, causal processing, camera-motion compensation, latency and failure recovery would also need evaluation.

These are proposed experiments, not results or commitments fulfilled by the present study. Their value is that each would resolve a specific inferential gap instead of merely increasing a development score.

\section{Conclusion}
\method{} provides a selective, inspectable implementation for a restricted monocular precession problem. A new 96-scene synthetic population produced 38 correct joint estimates, no wrong emissions, 57 withholding decisions and one incomplete execution in the primary pass. Scientific repetition was established for 95 of 96 scene pairs. The registered coverage and complete-execution requirements were not met.

The study supports an auditable case report with reproducible descriptive arithmetic and internal scientific repetition. It does not support state-of-the-art superiority, calibrated confidence, general identifiability or flight readiness. Its most transferable result is the need to preserve the distinction between a fit, an accepted estimate, a withheld answer and a computation that never established a valid answer.

\section*{Data, code and authoring statement}
The manuscript source package includes the per-attempt outcome/error table, the hash-pinned original score compressed with gzip, derivation and verification scripts, generated figures, bibliography, a literature ledger and an evidence map. It also contains the original protocol, attempted-case registration and frozen source excerpts for the measurement, fitting and decision computations. The exact scoring adapter, metric helpers, scientific comparator, comparison policy and completion seal are included for inspection. These materials reproduce the descriptive arithmetic and permit inspection of the stated implementation. A separate archive supplies all 96 original reference files, 192 reports and 192 execution receipts. Its verifier reproduces physical scoring from these originals; it does not rerun image inference. They do not contain the complete inference runtime or all rendered images, and no public data DOI is claimed. The arXiv source archive carries this material in its ancillary reproducibility package.

AI tools assisted software development, literature discovery, evidence inspection and drafting. Automated adversarial reviews were software-agent reviews, not human peer review. Carlos Eduardo Sanoja is the sole named author and retains scientific responsibility and the decision to submit.

\clearpage
\appendix
\section{Endpoint correction and its limitations}\label{app:geometry}
This derivation documents the objective used by the implementation. It is a reduction of orthogonal line fitting and constrained ray fitting; no priority claim is made for the underlying mathematics.

For $q_z\ne0$, define the finite vanishing point $v=q_{xy}/q_z$ and centered endpoints $x_j=p_j-v$. Let $n$ be a unit normal to a candidate line through $v$. The minimum squared correction of the two endpoints to that line is
\begin{equation}
 \min_{\|n\|=1}\sum_{j=1}^2(n^\top x_j)^2
 =\lambda_{\min}(x_1x_1^\top+x_2x_2^\top).
\end{equation}
Since $d_j=-q_zx_j$, the matrix in the eigenvalue expression is $S/q_z^2$. In two dimensions,
$\lambda_{\min}(S)=\det(S)/\lambda_{\max}(S)$, and
$\det(S)=\det[d_1,d_2]^2=q_z^2(\ell^\top q)^2$.
Substitution gives Equation~\eqref{eq:line} without subtracting nearly equal eigenvalues. Distinct endpoints and nonzero $q$ exclude the case $\lambda_{\max}(S)=0$. At $q_z=0$, $d_1=d_2=q_{xy}$, so
\begin{equation}
 d_{\rm line}^2=\frac{(\ell^\top q)^2}{2\|q_{xy}\|^2}.
\end{equation}
This equals the sum of endpoint corrections to their best line of fixed direction $q_{xy}$, with the line offset fitted through their midpoint.

For the common-ray constraint, use a unit direction $u$ and nonnegative endpoint distances $\alpha_1,\alpha_2$. The problem is
\begin{equation}
 \inf_{\|u\|=1,\ \alpha_j\geq0}
 \sum_j\|x_j-\alpha_ju\|^2.
\end{equation}
For fixed $u$, $\alpha_j=\max(0,u^\top x_j)$. If $x_1^\top x_2>0$, the principal direction can project both vectors with the same sign and the unconstrained line optimum is feasible. If $x_1^\top x_2\leq0$, the optimum has one endpoint at the origin of the ray; orienting toward the longer vector leaves squared cost $\min(\|x_1\|^2,\|x_2\|^2)$. Rescaling yields Equation~\eqref{eq:ray}. Equality of the dot product lies on the branch boundary; the formula remains continuous.

The nonnegative-distance problem includes the ray origin. For an actual segment with distinct finite-depth endpoints, attaining that origin can be physically inadmissible; the result is then only an infimum. This is why the objective alone cannot replace positive-ray support checks. It also cannot establish that the endpoints arose from the same physical generator, or that one pixel is an appropriate noise scale. Perspective incidence is exact under the model; its use as a statistical loss is an additional assumption.

\section{Frozen numerical search envelope}\label{app:numerics}
These controls describe the evaluated source, not recommended general-purpose defaults. Pixel preparation subtracts the first source timestamp, and the physical runtime requires a zero-based source calendar. Let $T=t_{F-1}$ be the resulting full-source span and $\Delta T_{\rm TRAIN}$ the span of source timestamps in the declared TRAIN index set. In an end-block fold these spans can differ.

\subsection{Cone bank}
The 480 starts use 240 uniformly spaced initial frequencies and two depth lifts. Frequency is bounded by
\begin{equation}
 \max(0.5,360/\Delta T_{\rm TRAIN})\leq\Omega\leq30
 \quad\text{degrees/s}.
\end{equation}
The aperture lies in $[10^{-6},\pi-10^{-6}]$ radians, while the three rotation-vector coordinates are unbounded. Frequency is scaled by $360/\Delta T_{\rm TRAIN}$. Each fit sets the L-BFGS-B iteration and objective/gradient-bundle limits to 350, with a line-search limit of 30, gradient tolerance $10^{-6}$ and function tolerance $10^{-12}$. The finite-difference step is $10^{-5}$ in the scaled coordinates, with one-sided differences at bounds. A bundle includes derivative work; these are configured solver limits, and 350 is not a hard bound on all scalar loss callbacks. Line-search and finite-difference work are recorded separately.

The cone numerical-termination gate accepts either the solver success flag or a nonnegative canonical loss at most $10^{-20}$. This floor rule is not a stationarity or accuracy certificate. Selected triaxial fits instead retain their solver success flag. These implementation-specific meanings qualify the word ``converged'' in gate names.

\subsection{Triaxial bank}
Up to eight diverse cone parents supply initializations. The seed construction explores $J\in\{2,4,8\}$, $d\in\{0.2,0.8\}$, both regimes, their specified signs and four phases per regime period; the frozen ranking/diversity procedure chooses 128 starts. The continuous energy fraction is bounded by $[10^{-5},1-10^{-5}]$, $J$ by $[1.51,12]$, $d$ by $[0.001,0.999]$ and phase fraction by $[-3,3]$. Rotation-vector coordinates are unbounded. The scale $r$ satisfies
\begin{equation}
 \max(0.5,360/T)\leq r\leq30\quad\text{degrees/s},
 \qquad \|H\|=(\pi/180)rJ .
\end{equation}
Thus, the bound on $r$ is not a bound of 30 degrees/s on the reported mean precession of every triaxial trajectory. The eight parameter scales are $(1,1,1,0.3,360/T,3,0.5,0.5)$.

SciPy's trust-region reflective least-squares solver first uses a 350-evaluation budget and function, step and gradient tolerances of $10^{-9}$. After rotation-vector canonicalization, a second 350-evaluation stage uses $10^{-10}$ tolerances. If the second stage reports nonconvergence, a predefined local refinement allows 1000 evaluations at the latter tolerances and is adopted only if its raw cost is no greater than the previous cost plus $10^{-12}\max(1,\text{previous cost})$. These are solver-counter budgets, not total scalar residual-call or image-processing budgets; numerical Jacobian evaluations can require additional residual calls. They do not guarantee convergence. The fixed internal refinement belongs to each original attempt; it is distinct from replacing or rerunning a scene after its outcome is known. Frozen source excerpts expose the exact seed ordering, cost definitions and callback accounting.

\clearpage
\section{Gate inventory and decision semantics}\label{app:gates}
Table~\ref{tab:gates} lists all 27 gates in the evaluated decision policy. Names are shortened for readability; their exact machine-readable names and the family-selection policy accompany the source package. A full-fit rejection can terminate before temporal searches, but an emitted point requires every applicable check. A missing diagnostic is not a negative measurement and cannot be converted into a valid withholding decision without the runtime establishing that contract.

\begin{longtable}{r p{54mm} p{82mm}}
\caption{Complete gate inventory, grouped by execution stage.}\label{tab:gates}\\
\toprule No. & Gate & Role\\\midrule
\endfirsthead
\toprule No. & Gate & Role\\\midrule
\endhead
1 & Resolved rate & Rate resolution under the frozen engineering rule\\
2 & Source sampling & Acquisition sampling support\\
3 & Resolved cycle & Sufficient resolved temporal cycle\\
4 & Camera alignment & Supported camera geometry\\
5 & Cone aperture & Nondegenerate axis excursion\\
6 & Phase coverage & Admitted phase support\\
7 & Full winner converged & Selected minimum's numerical-termination rule\\
8 & Full pixel support & Selected fit's training endpoint support\\
9 & Full angular support & Selected fit's training direction support\\
10 & Full positive-ray support & Strong selected generator-ray support\\
11 & Full quadrature converged & Triaxial path refinement below $10^{-3}$; explicitly not applicable to analytic cone\\
12 & Full numerical screen & Local numerical parameter screening\\
13 & Eight complete folds & All eight actual temporal contexts available\\
14 & Nested training converged & Every selected fold satisfies its numerical-termination rule\\
15 & Nested line quality & Pixel and angular training checks in all folds\\
16 & Nested parameter stability & Fold/full linked points not materially distinct\\
17 & Nested quadrature converged & Refinement rule in each applicable fold\\
18 & Nested numerical screen & Parameter screen in each fold\\
19 & Nested training rays & Strong positive-ray support in each training context\\
20 & Nonempty held-out blocks & Held-out line evidence in every block\\
21 & Pooled held-out pixels & Endpoint quality across held-out predictions\\
22 & Pooled held-out angles & Direction quality across held-out predictions\\
23 & Held-out rays & Positive-ray support in every held-out block\\
24 & Conditional axis radius & Computed stability radius below $10^\circ$\\
25 & Conditional rate radius & Computed stability radius below 15\%\\
26 & Distinct rivals worse & Training-compatible distinct alternatives worse on held-out data\\
27 & Rival optimization & Nested rivals satisfy declared optimizer rules\\
\bottomrule
\end{longtable}

\subsection{Conditional dispersion and rival contrasts}\label{app:decisiondiagnostics}
After all $m_f=8$ folds return valid linked targets, the nominal target is $(\hat h_0,\hat\Omega_0)$ and the fold targets are $(\hat h_j,\hat\Omega_j)$, $j=1,\ldots,m_f$. Every rate is evaluated over the same full source calendar. Each unit fold axis is sign-aligned with the nominal axis:
\begin{equation}
 \tilde h_j=\underset{x\in\{\hat h_j,-\hat h_j\}}{\operatorname{argmax}}
 x^\top\hat h_0,\qquad
 \bar h_b=\frac1{m_f}\sum_j\tilde h_j,\qquad
 \bar\Omega_b=\frac1{m_f}\sum_j\hat\Omega_j.
\end{equation}
The vector mean $\bar h_b$ is not renormalized. The implementation forms the block-dispersion quantities
\begin{align}
 C_h&=\frac{m_f-1}{m_f}\sum_j
       (\tilde h_j-\bar h_b)(\tilde h_j-\bar h_b)^\top,
 &s_h&=\sqrt{\max(0,\lambda_{\max}(C_h))},\\
 s_\Omega^2&=\frac{m_f-1}{m_f}\sum_j
       (\hat\Omega_j-\bar\Omega_b)^2.
\end{align}
Let $t_{\nu,p}$ denote the $p$ quantile of Student's $t$ distribution with $\nu$ degrees of freedom. With the fixed nominal multiplier parameter $\alpha_{\rm nom}=0.01$, the reported diagnostics are
\begin{equation}
 R_h=\frac{180}{\pi}\sqrt{2}\,
 t_{m_f-1,\,1-\alpha_{\rm nom}/4}\,s_h,
 \qquad
 R_\Omega=100\,
 \frac{t_{m_f-1,\,1-\alpha_{\rm nom}/2}\,s_\Omega}{\hat\Omega_0}.
\end{equation}
The gates require $R_h<10$ degrees and $R_\Omega<15\%$. The axis calculation converts ambient-vector dispersion into degrees; it is not an exact spherical confidence construction. The overlapping training sets and shared upstream features make the fold estimates correlated. In particular, $\alpha_{\rm nom}=0.01$ is a fixed multiplier choice, not evidence of calibrated 99\% coverage. These quantities are descriptive conditional stability diagnostics.

In each fold, let $w$ denote its TRAIN-selected linked point and $r$ a retained, materially distinct rival marked TRAIN-compatible by the frozen quality rules. Canonical training and held-out losses use the cost normalization in Equation~\eqref{eq:loss}. Every such rival must satisfy
\begin{equation}
 L_r^{\rm held}-L_w^{\rm held}>0,
 \qquad
 c_r\ \lor\
 \left[L_r^{\rm TRAIN}\geq9\max(L_w^{\rm TRAIN},10^{-12})\right],
\end{equation}
where $c_r$ is the rival's declared numerical-termination flag under Appendix~\ref{app:numerics}. Thus a nonconverged rival is permitted by this rule only when its training loss meets or exceeds the stated ninefold threshold. The held-out comparison has no ratio margin, effect-size threshold or population $p$-value. Both conditions must hold for every required rival in every fold. If there is no such retained rival, the universal check passes vacuously; this means only that none was found in the finite search. Missing or malformed required held-out evidence produces an incomplete comparison rather than a passing contrast.

The dimension penalty does not make frames independent. If two families have losses $L_T,L_C$, their selection-score difference is
\begin{equation}
 \Delta S=N\log(L_T/L_C)+3\log N
\end{equation}
away from the loss floor. Duplicating every frame $r$ times leaves the mean losses unchanged but changes this difference to
$rN\log(L_T/L_C)+3\log(rN)$. The selected family can change although no new independent information has been acquired. This simple consequence is one reason the penalty is described as a heuristic and $N$ is not called an effective sample size.

\section{Evidence identity and reproduction boundaries}\label{app:provenance}
The final score and original protocol are distinct artifacts. The following identities bind the closed result used in this manuscript:

\noindent\textbf{Descriptive score, SHA-256}\\
\hash{8de4a733955f5c721f5f81a1321c6ccd1fb92e4f55565e27331fd40d6063a98f}\par
\noindent\textbf{Descriptive supplement completion, SHA-256}\\
\hash{810f4009df3d0436272f3a984ea0be8407ab2f455a4149a5e8b7aaa397868dd4}\par
\noindent\textbf{Original reproduction completion, SHA-256}\\
\hash{9454e344c8728f61bcef7d50445d29b9d79024296446cbeafe9a41d381f16aa7}\par
\noindent\textbf{Inference OCI image identity, SHA-256}\\
\hash{82121b44e7d1cc2051f30642a8ff8fbf1ae9b4c5c8d64fed5f0eac4cb38a7e39}\par
\noindent\textbf{Registered generator content identity, SHA-256}\\
\hash{c0bb0a6560d56b36832220af0b74469264da2a4a20eb8d96eb750e31737e4adb}\par
Hashes support identity checks only when the corresponding artifacts are available. They are not proof of scientific correctness or substitutes for an accessible archive. The original descriptive score is included as a gzip file; decompression recovers the exact bytes bound by the first digest above. The compact CSV has one row for every scene and repetition; missing point errors for withheld or incomplete outcomes remain empty, not zero. Its source digest is checked by the derivation script before extracting the rows from the original score.

The source package separates four reproduction levels:
\begin{enumerate}
\item \textbf{Manuscript build.} Compile the supplied \LaTeX{} and bibliography using the included figures and tables. This verifies the document, not the experiment.
\item \textbf{Descriptive arithmetic.} Run the supplied Python script on the compact CSV or the included compressed original score. This rebuilds counts, conditional-error summaries and plots without inference. A separate verifier checks all 192 CSV rows against the original score, source identities, citation coverage and elementary geometry controls.
\item \textbf{Physical metric reproduction.} The companion original-input archive contains 96 reference files, 192 original reports and 192 execution receipts. A hash-bound verifier extracts the archived reference targets and recomputes sign-only axis and percentage rate errors using the frozen metric functions. All 76 emitted-attempt error pairs reproduced the archived floating-point values exactly in the checked environment; all 192 outcome labels agree. The separate 192-row point export preserves unavailable values and links back to the originals. This post-closure check is a reproduction of scoring, not a new accuracy population or image-inference run.
\item \textbf{Image-to-report reproduction.} Obtain the pinned runtime, frozen candidate, 96-scene inputs and evaluation artifacts. The compact package alone cannot perform this level. A new execution must preserve the original attempted-case manifest and comparison policy, and must report failures rather than selecting successful runs.
\end{enumerate}
The project evidence map supplies repository-relative locations for the protocol, generator, frozen measurement code, failure record, historical corrections and numerical control. A separate source map binds the exact excerpts distributed with the package. Some original project locations remain local; the maps do not imply that every historical archive is included. No access key or authenticated link token is embedded in the manuscript or source package.

\clearpage
\section{Claim boundaries for reading and reuse}
\begin{table}[!ht]
\centering\small
\caption{What the evidence supports and what remains unestablished.}
\begin{tabularx}{\linewidth}{YY}
\toprule
Supported within this study & Not established\\
\midrule
38 correct linked estimates on the primary synthetic pass & Error probability of zero, or a calibrated per-scene confidence\\
25/64 correct supported coverage & Registered minimum coverage or general target recovery\\
95 scientifically equivalent scene pairs & 96/96 repetition, cross-platform equality or independent-laboratory reproduction\\
No reference data mounted in inference & Independence of all generator and estimator assumptions\\
Physical cone and torque-free triaxial trajectories & Correct physical identity of every extracted edge\\
Complete prescribed search banks retained & Exhaustive search or global identifiability\\
Cold temporal numerical searches & Pixel-isolated end-to-end holdout\\
Correct estimates on some stress cases & A validated expansion beyond the stated supported domain\\
Tiny SSE decreases in a separate numerical control & Improved physical accuracy or coverage\\
\bottomrule
\end{tabularx}
\end{table}

\FloatBarrier
\bibliographystyle{unsrtnat}
\begingroup\raggedright
\bibliography{references}
\endgroup
\end{document}